\documentclass{llncs}
\usepackage[utf8]{inputenc}
\usepackage{soul}
\usepackage{biblatex}
\usepackage{graphicx}
\usepackage{hyperref}
\usepackage{comment}
\usepackage[colorinlistoftodos]{todonotes}
\usepackage{xspace}
\usepackage{booktabs} 
\usepackage{array} 
\usepackage{anyfontsize}
\usepackage{amsmath}
\usepackage{cleveref}[2012/02/15]
\usepackage{subcaption}
\usepackage{float}
\crefformat{footnote}{#2\footnotemark[#1]#3}

\newcommand{\DMNp}{cDMN\xspace
}

\definecolor{inputgreen}{RGB}{182, 215, 168}
\definecolor{inputblue}{RGB}{207, 226, 243}
\definecolor{glossaryorange}{RGB}{255, 123, 89}
\definecolor{executered}{RGB}{236, 155, 164}

\newcommand{\chalname}[1]{\textit{#1}}
\newcommand{\firstval}[0]{\cellcolor[gray]{0.9} }
\newcommand{\secondval}[0]{\cellcolor[gray]{0.7} }
\newcommand{\thirdval}[0]{\cellcolor[gray]{0.6} }
\newcommand{\fourthval}[0]{\cellcolor[gray]{0.4} }
\newcommand{\emptyval}[0]{\cellcolor[gray]{1.0} }
\newcommand{\inputcol}[0]{\cellcolor{inputgreen} }
\newcommand{\outputcol}[0]{\cellcolor{inputblue} }
\newcommand{\glossarycol}[0]{\cellcolor{glossaryorange}}
\newcommand{\executecol}[0]{\cellcolor{executered}}
\newcommand{\dmnfont}[0]{\fontsize{7}{9} \fontfamily{put} \selectfont \setlength{\belowrulesep}{0pt}}
\usepackage{adjustbox}
\usepackage{array}

\newcolumntype{R}[2]{%
    >{\adjustbox{angle=#1,lap=\width-(#2)}\bgroup}%
    l%
    <{\egroup}%
}
\newcommand*\rot{\multicolumn{1}{R{0}{1em}}}

\usepackage[table]{colortbl}

\makeatletter
\def\blfootnote{\xdef\@thefnmark{}\@footnotetext}
\makeatother

\begin{document}

\title{Tackling the DMN Challenges with \DMNp: \newline A Tight Integration of DMN and Constraint Reasoning}
\author{Bram Aerts, Simon Vandevelde, Joost Vennekens}
\institute{
KU Leuven, De Nayer Campus, Dept. of Computer Science \\
\email{\{b.aerts, s.vandevelde, joost.vennekens\}@kuleuven.be}}

\maketitle
\vspace{-1em}

\section*{Abstract}
\vspace{-0.15em}

This paper describes an extension to the DMN standard, called cDMN. 
It aims to enlarge the expressivity of DMN in order to solve more complex problems, while retaining DMN’s goal of being readable by domain experts. 
We test cDMN by solving the most complex challenges posted on the DM Community website. 
We compare our own cDMN solutions to the solutions that have been submitted to the website and find that our approach is competitive, both in readability and compactness. 
Moreover, cDMN is able to solve more challenges than any other approach.

\vspace{-0.2em}

\section{Introduction}
\vspace{-0.2em}

The Decision Model and Notation (DMN) \cite{DMN} standard, designed by the Object Management Group (OMG), is a way of representing data and decision logic in a readable, table-based way.
It is intended to be used directly by business experts without the help of computer scientists.

While DMN is very effective in modeling deterministic decision processes, it lacks the ability to represent more complex kinds of knowledge.
In order to explore the boundaries of DMN, the Decision Management Community website\footnote{\url{https://dmcommunity.org/}} issues a monthly decision modeling challenge.
Community members can then submit a solution, using their preferred decision modeling tools or programming languages.
This allows solutions for complex problems to be found and compared across multiple DMN-like representations.
So far, none of the available solvers have been able to solve all challenges. 
Moreover, the available solutions sometimes fail to meet the readability goals of DMN, because the representation is either too complex or too large.

\blfootnote{This research received funding from the Flemish Government under the “Onderzoeksprogramma Artificiële Intelligentie (AI) Vlaanderen” programme.}


In this paper, we propose an extension to the DMN standard, called \DMNp.
It allows more complex problems to be solved, while remaining readable by business users. 
The main features of \DMNp are constraint modeling, quantification, and the use of concepts such as types and functions.
We test the usability of \DMNp  on the decision modeling challenges.

In \cite{DeryckM.2019Actt}, we presented a preliminary version of constraint modeling in DMN. In the current paper, we extend this by adding quantification, types, functions, relations, data tables, optimization and by evaluating the formalism on the DMN challenges.

The paper is structured as follows. 
In section \ref{sec:prelim} we briefly describe the DMN standard. 
Section \ref{sec:chaloverview} gives an overview of the challenges used in this paper. 
After this, we touch on the related work in section \ref{sec:relatedwork}.
We discuss both syntax and semantics of our new notation in section \ref{sec:cDMN}.
Section \ref{sec:implementation} briefly discusses the implementation of our cDMN solver.
We compare our notation with other notations and evaluate its added value in section \ref{sec:results}, and conclude in section \ref{sec:conclusion}.

\vspace{-0.2em}
\section {Preliminaries: DMN}\label{sec:prelim}
\vspace{-0.2em}

\begin{figure}
    \vspace{-1em}
    \includegraphics[width=\linewidth]{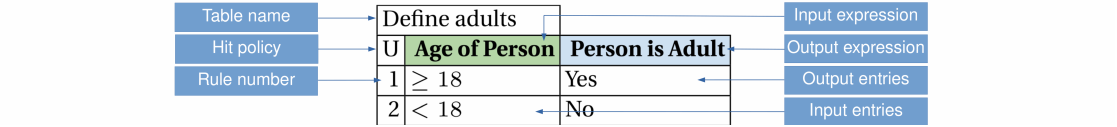}
    \vspace{-2em}

    \caption{Decision table to define whether a person is an adult.}
    \label{fig:decisiontable}
    \vspace{-1.5em}
\end{figure}

DMN consists of two components: a Decision Requirements Diagram (DRD), which is a graph that expresses the structure of the model, and Decision Tables, which contain the in-depth business logic. 
An example of such a decision table can be found in Figure \ref{fig:decisiontable}. 
It consists of a number of input columns (darker green) and a single output column (lighter blue). 
Each row is read as: if the input conditions are met (e.g., if ``Age of Person'' satisfies the comparison ``$\geq$ 18'' ), then the output expression is assigned the value of the output entry (e.g. ``Person is Adult'' is assigned value ``Yes''). 
Only single values, such as strings and numbers, can be used as output entries. 
In the case where no row matches the input, then each output is either set to the special value \textit{null} (which is typically taken to indicate an error in the specification) or to the output's default value, if one was provided.

The behaviour of a decision table is determined by its hit policy.
There are a number of \emph{single hit} policies, which define that a table can have at most one output for each possible input, such as ``Unique'' (no overlap may occur), ``Any'' (if there is an overlap, the outputs must be the same) and ``First'' (if there is an overlap, the first applicable row should be selected). There exist also \emph{multiple hit} policies such as C (collect the output of all applicable rows in a list) and C+ (sum the output of all applicable rows).
Regardless of which hit policy is used, each decision table uniquely determines the value of its output(s).

The entries in a decision table are typically written in the (Simple) Friendly Enough Expression Language, or (S-)FEEL, which is also part of the DMN standard. 
S-FEEL allows to express simple values, lists of values, numerical comparisons, ranges of values and calculations. 
Decision tables with S-FEEL are generally considered quite readable by domain experts.

In addition, DMN also allows more complex FEEL statements in combination with boxed expressions, as will be illustrated in Figure \ref{fig:FEELmap}.
However, this also greatly increases complexity of the representation, which makes it unsuitable to be used by domain experts without the aid of knowledge experts.

\vspace{-0.15em}

\section{Challenges Overview}\label{sec:chaloverview}
\vspace{-0.15em}

Of all the challenges on the DM Community website, we selected those that did not have a straightforward DMN-like solution submitted.
The list of the 21 challenges that meet this criterion can be found in the cDMN documentation\footnote{\url{https://cdmn.readthedocs.io/en/latest/community.html}}.

We categorize these challenges according to four different properties.
Table \ref{table:problemproperties} shows the list of properties, and the percentage of challenges that have this property.

The most frequent property is the need for aggregates (57.14\%), such as counting the number of violated constraints in \chalname{Map Coloring with Violations} or summing the number of calories of ingredients in \chalname{Make a Good Burger}. 
The second most frequent property is having constraints in the problem description (33.33\%). 
E.g., in \chalname{Who Killed Agatha}, the killer hates the victim and is no richer than her; or the constraint in \chalname{Map Coloring} states that two bordering countries can not share the same color.
The next property, universal quantification (28.75\%), is that a statement applies to every element of a type, for example in \chalname{Who Killed Agatha?}: nobody hates everyone.
The final property, optimization, occurs in 23.81\% of the challenges.
For example, in \chalname{Zoo, Buses and Kids} the cheapest set of buses must be found.



\begin{table}
    \vspace{-1.1em}
    
    \centering
    \begin{tabular}{l|r}
        Property & (\%) \\
        \hline
        1. Aggregates needed        & 57.14  \\
        2. Constraints              & 33.33  \\
        3. Universal quantification & 28.75  \\
        4. Optimization             & 23.81  \\
    \end{tabular}
    \caption{Percentage of occurrence of properties in challenges.}
    \label{table:problemproperties}
    \vspace{-4em}
\end{table}

\vspace{-0.15em}

\section{Related Work} \label{sec:relatedwork}
\vspace{-0.15em}

It has been recognized that even though DMN has many advantages, it is somewhat limited in expressivity \cite{calvanese2017semantic, DeryckM.2019Actt}.
This holds especially for decision tables with S-FEEL, the fragment of FEEL that is considered most readable.
While full FEEL is more expressive, it is not suitable to be used by domain experts without the aid of knowledge experts. 
Moreover, it does not provide a solution to other shortcomings, such as the lack of constraint reasoning and optimization.

One of the systems that does effectively support constraint solving in a readable DMN-like representation is the OpenRules system \cite{OpenRules}. 
It enables users to define constraints over the solution space by allowing ``Solver Tables'' to be added alongside decision tables. 
In contrast to standard decisions, which assign a specific value to an output, Solver Tables allow for setting constraints on the output space.
OpenRules offers a number of \textit{DecisionTableSolve-Templates}, which can be used to specify these constraints.
It is possible to either use these predefined templates, or define such a template manually if the predefined ones are not expressive enough. 
Even though this system extends the range of applications that can be handled, there are three reasons why it does not offer the ease of use for business users that we are after. 
First, because of the wide range of available templates for solver tables, which differ from that of standard decision tables, using the OpenRules constraint solver entails a steep learning curve.
Second, the solver's functionality can only be accessed through the Java API, which goes against the DMN philosophy \cite[p.~13]{DMN}.
Third, because of the lack of quantification in OpenRules, solutions are generally not independent of domain size, which reduces readability.


Another system that aims to increase expressiveness of DMN is Corticon \cite{Corticon}. 
It implements a basic form of constraint solving by allowing the user to filter the solution space.
While this approach indeed improves expressiveness, it decreases readability.
Moreover, some constraints can only be expressed by combining a number of rules and a number of filters. 
For example, when expressing ``all female monkeys are older than 10 years'', this is be split up in two parts; (1) a rule that states \texttt{Monkey.gender = female \& Monkey.Age < 10 THEN Monkey.illegal = True} and (2) a filter that states that a monkey cannot be illegal: \texttt{Monkey.illegal = False}.
There are no clear guidelines about which part of the constraints should be in the filter and what should be a rule.

In \cite{calvanese2017semantic}, Calvanese et al propose an extension to DMN which allows for expressing additional domain knowledge in Description Logic. They share our goal of extending DMN to express more complex real-life problems. However, they introduce a completely separate Description Logic formalism, which seems too complex for a domain expert to use.
Unfortunately, they did not submit any solutions to the DMN Challenges, which leaves us unable to compare its expressiveness in practice.

\vspace{-0.2em}

\section{\DMNp: syntax \& semantics} \label{sec:cDMN}
\vspace{-0.2em}

While DMN allows users to elegantly represent a deterministic decision process, it lacks the ability to specify constraints on the solution space.
The \DMNp framework extends DMN, by allowing constraints to be represented in a straightforward and readable manner.
It also allows for representations that are independent of domain size by supporting types, functions, relations and quantification.

We now explain both the usage and the syntax of every kind of table present in \DMNp.

\subsection{Glossary}\label{sub:glossary}

In logical terms, the ``variables'' used in standard DMN correspond to constants (i.e., 0-ary functions).
\DMNp extends these by adding n-ary functions and n-ary relations.
Similarly to OpenRules and Corticon, we allow the user to define their vocabulary by means of a glossary.
This glossary contains every symbol used in a \DMNp model. 
It consists of at most five glossary tables, which each enumerate a different kind of symbol.
An example glossary for the \chalname{Who Killed Agatha?} challenge is given in Figure \ref{tab:glossary_example}.

\begin{figure}[h]
    \vspace{-2em}
    \centering
    {
    \dmnfont
    \begin{tabular}{|l|l|l|l|}
        \cmidrule{1-3}
         \multicolumn{3}{|c|}{\textbf{Type} \glossarycol} \\
        \hline
        \textbf{Name} & \textbf{Type} & \textbf{Values}\\
        \hline
        Person &  string & Agatha, Butler, Charles\\
        \hline
        Number & int & [0..100]\\
        \hline
    \end{tabular}
    \quad
    \begin{tabular}{|l|}
        \cmidrule{1-1}
         \multicolumn{1}{|c|}{\textbf{Relation} \glossarycol} \\
        \hline
        \textbf{Name}\\
        \hline
        Person hates Person\\
        \hline
        Person is richer than Person  \\
        \hline
    \end{tabular}
    
    \begin{tabular}{|l|}
        \cmidrule{1-1}
        \multicolumn{1}{|c|}{\textbf{Boolean} \glossarycol} \\
        \hline
        \textbf{Name}\\
        \hline
        Suicide \\
        \hline
    \end{tabular}
    \quad
    \begin{tabular}{|l|l|}
        \cmidrule{1-2}
         \multicolumn{2}{|c|}{\textbf{Constant} \glossarycol} \\
        \hline
        \textbf{Name} & \textbf{Type}\\
        \hline
        Killer & Person\\
        \hline
    \end{tabular}
    \quad
    \begin{tabular}{|l|l|}
        \cmidrule{1-2}
         \multicolumn{2}{|c|}{\textbf{Function} \glossarycol} \\
        \hline
        \textbf{Name} & \textbf{Type}\\
        \hline
        Hatees of Person & Number\\
        \hline
    \end{tabular}

    }
    \caption{An example \DMNp glossary for the \chalname{Who Killed Agatha?} problem.}
    \label{tab:glossary_example}
    \vspace{-2em}
\end{figure}

In the \emph{Type} table, \emph{type} symbols are declared. 
The value of each type is a set of domain elements, specified either in the glossary or in a data table (see \ref{sec:datatable}).
An example is the type \texttt{Person}, which contains the names of people.

In the \emph{Function} table, a symbol can be declared as a {\em function} of one or more types to another. 
The infix operator \texttt{of} is used to apply the function to its argument(s).
For example, the \texttt{Hatees of Person} function denotes how many people a person hates. 
It maps each element oftype \texttt{Person} to an element of type \texttt{Number}. 
Functions with $n > 1$ arguments can be declared by separating the \textit{n} arguments by the keyword \texttt{and}.

For each domain element, a constant with the same name is automatically introduced, which allows the user to refer to this domain element in constraint or decision tables.
For instance, the user can use the constant \texttt{Agatha} to refer to the domain element \texttt{Agatha}.
In addition, the \emph{Constant} table allows other constants to be introduced.
Recall that such logical constants correspond to standard DMN variables.
In our example case, we use a constant \texttt{Killer} of the type \texttt{Person}, which means it can refer to any of the domain elements \texttt{Agatha}, \texttt{Butler} or \texttt{Charles}.

In the \emph{Relation} table, a verb phrase can be declared as a {\em relation} on one or more given types.
For instance, the relation \texttt{Person is Adult} denotes for each \texttt{Person} whether they are an adult.
n-ary predicates can be defined by using \textit{n} arguments in the name, e.g. \texttt{Person is richer than Person} is a relation with two arguments (both of the type \texttt{Person}), that denotes whether one person is richer than another.

The \emph{Boolean} table contains \emph{boolean} symbols (i.e. propositions), which are either true or false.
An example is the boolean \texttt{Suicide}, which denotes whether the murder is a suicide.

\subsection{Decision Tables and Constraint Tables}\label{sub:decision_constraint}


As stated earlier in Section \ref{sec:prelim}, a standard decision table uniquely defines the value of its outputs.
We extend DMN by allowing a new kind of table, called a {\em constraint table}, which does not have this property.

Whereas decision tables only allow single values to appear in output columns, our constraint tables allow arbitrary S-FEEL expressions in output columns, instead of only single values.
Each row of a constraint table represents a logical {\em implication}, in the sense that, if the conditions on the inputs are satisfied, then the conditions on the outputs must also be satisfied.
This means that if, for instance, none of the rows are applicable, the outputs can take on an arbitrary value, as opposed to being forced to {\em null}.
In constraint tables, no default values can be assigned.
Because of these changes, a set of \DMNp tables does not define a single solution, but rather a solution space containing a set of possible solutions.

We introduce a new \textit{hit policy} to identify constraint tables. We call this the {\em Every} hit policy, denoted as {\em E*}, because it expresses that every implication in the table must be satisfied. 
An example of this can be found in Figure \ref{tab:agatha}, which states that each person hates less than 3 people.   

\DMNp does not only introduce constraint tables, it also extends the expressions that are allowed in column headers, both in decision and constraint tables.
Such a header can consist of the following expressions:
(1) a type $Type$; 
(2) an expression of the form ``$Type$ called $name$''; 
(3) a constant;
(4) an expression of the form ``$Function$ of $arg_1$ and ... and $arg_n$'', where each of the $arg_i$ is  
another header expression;
(5) an arithmetic combination of header expressions (such as a sum).

The first two kinds of expressions are called {\em variable-introducing} header expressions. 
They allow \textit{universal quantification} in \DMNp.
An input column whose header consists of such a variable-introducing expression is called a {\em variable-introducing} column. 
Each such column introduces a new universally quantified variable. Subsequent uses of the same type name (in case of the first kind of variable-introducing expression) or of the variable name (in case of the second kind) then refer back to this universally quantified variable.
Whenever a type or variable name appears in a header of a column that is itself not variable-introducing, a unique preceding variable-introducing column that has introduced this variable must exists.

The table in Figure \ref{tab:agatha} shows an example of quantification in \DMNp. 
It introduces a universally quantified variable of the type \textit{Person}, stating that every person hates less than three others.
To illustrate the use of named variables, Figure \ref{tab:map} defines variables \texttt{c1} and \texttt{c2}, both of the type \texttt{Country}, and states that when those countries are bordering, they cannot have the same color.



In summary, this section has discussed three ways in which \DMNp extends DMN.
First, the hit policy \textit{E*} changes the semantics of the table.
Second, constraint tables allow S-FEEL expressions in the output columns.
Third, \DMNp allows quantification, functions, predicates and calculations to be used in both decision tables and constraint tables.

\begin{figure}[h]
    \vspace{-2em}
    \centering
    {
    \dmnfont
    \begin{tabular}{|r|l|l|}
        \cmidrule{1-2}
        \multicolumn{2}{|l|}{Noone hates all} & \multicolumn{1}{c}{}\\
        \hline
        E* & \inputcol \textbf{Person} & \outputcol \textbf{Hatees of Person}\\
        \hline
        1 & - & \textless 3 \\
        \hline
    \end{tabular}
    }
    \caption{Part of the implementation of ``Nobody hates everyone'' in \chalname{Who Killed Agatha?}.}
    \label{tab:agatha}
    \vspace{-2em}
\end{figure}

\begin{figure}[h]
    \centering
    {
    \dmnfont
    \begin{tabular}{|r|l|l|l|l|}
        \cmidrule{1-4}
        \multicolumn{4}{|l|}{Bordering countries can not share colors} & \multicolumn{1}{c}{}\\
        \hline
        E* & \inputcol \textbf{Country called c1} & \inputcol \textbf{Country called c2}
           & \inputcol \textbf{c1 and c2 are Bordering} & \outputcol \textbf{Color of c1}\\
        \hline
        1 & - & - & Yes & Not(Color of c2)\\
        \hline
    \end{tabular}
    }
    \caption{Example of a constraint table with quantification in \DMNp, defining that bordering countries can't share colors.}
    \label{tab:map}
    \vspace{-1.5em}
\end{figure}

\subsection{Data Tables} \label{sec:datatable}

Typically, problems can be split up into two parts: (1) the general logic of the problem, and (2) the specific problem instance that needs to be solved. 
Take for example the map coloring problem: the general logic consists of the rule that two bordering countries cannot share a color, whereas the instance of the problem is the specific map (e.g. Western Europe) to color.
\DMNp extends the DMN standard to include \textit{data tables}, which are used to represent the problem instances, separating them from the general logic.
The format of a data table closely resembles that of a  decision table, with a couple of exceptions.
Instead of a hit policy, a data table has ``data table'' in its name.
Furthermore, only basic values (integers, floats and domain elements) are allowed in data tables.
As an example, a snippet of the data table for the \chalname{Map Coloring} challenge is shown in Figure \ref{tab:datatable_example}.
This use of data tables offers several advantages.

\vspace{-0.5em}
\begin{enumerate}
    \item There is a methodological advantage: by separating the data tables from the decision tables, reusing the specification becomes easier.
    \item If the user chooses to enumerate the domain of a type in the glossary, then the system checks that each value is a data table indeed belongs to the domain of the appropriate type. 
    This helps to prevent errors or typos in the input data or glossary.
    If the user chooses not to enumerate a type in the glossary, then the type's domain defaults to the set of all values in the data table. 
    \item The cDMN solver is able to compute solutions faster, due to a different internal representation between data tables and decision tables.
\end{enumerate}
\vspace{-0.5em}

\begin{figure}[h]
    \vspace{-1em}
    \centering
    {
    \dmnfont
    \begin{tabular}{|r|l|l|l|}
        \cmidrule{1-3}
        \multicolumn{3}{|l|}{Data Table: Declaring which countries border} & \multicolumn{1}{c}{}\\
        \hline
         & \inputcol \textbf{Country called c1} & \inputcol \textbf{Country called c2}
            & \outputcol \textbf{c1 and c2 are Bordering}\\
        \hline
        1 & Belgium & France, Luxembourg, Netherlands, Germany & Yes \\
        \hline
        2 & Germany & France, Denmark, Luxembourg & Yes \\
        \hline
    \end{tabular}
    }
    \caption{Data table describing countries and their neighbours}
    \label{tab:datatable_example}
    \vspace{-4em}
\end{figure}

\subsection{Execute Table}

A standard DMN model defines a deterministic decision procedure.
It is typically always used in the same way: the external inputs are supplied by the user, after which the values of the output variables are computed by forward propagation.

In \DMNp, this is no longer the case.
We can fill in as many or as few variables as we want, and use the model to derive useful information about the not-yet-known variables.
By employing an \textit{execute table}, users can specify what the model is to be used for: model expansion or optimization.
Model expansion creates a given number of solutions, and optimization looks for the solution with either the lowest or heighest value for a given term.

\vspace{-0.5em}
\subsection{Semantics of \DMNp}

We describe the semantics of \DMNp by translating it to the FO($\cdot$) language used by the IDP system \cite{BogaertsBart2018PLaa}.
FO($\cdot$) is a rich extension of First Order Logic, adding concepts such as types, aggregates and inductive definitions.
The semantics of \DMNp is defined by the semantics of each of its sub-components.

It is straightforward to translate the glossary into an FO($\cdot$) vocabulary:
types, functions, constants, relations and booleans are each translated to their FO($\cdot$) counterpart.

Decision tables retain their usual semantics as described by Calvanese \cite{calvanese2017semantic}. 
We briefly recall this semantics. 
Each cell of a decision table $(i, j)$ corresponds to a formula $F_{ij}(x)$ in one free variable $x$. 
For instance, a cell ``$\leq 50$'' corresponds to the formula ``$x \leq 50$''.
A decision table with rows $R$ and columns $C$ is a disjunction of conjunctions, ${\bigvee_{i \in R} \bigwedge_{j \in C} F_{ij}(H_j)}$,
where $H_j$ is the header of column $j$. 
For example, the table in Figure \ref{fig:decisiontable} corresponds to the logical formula $(Age Of Person \geq 18 \land Person Is Adult = Yes) \lor  (Age Of Person < 18 \land Person Is Adult = No)$.

Data tables are simply a specific case of decision tables.

In \cite{DeryckM.2019Actt}, we defined the semantics of simple constraint tables (without quantification and functions) as a conjunction of implications:  
\newcommand{\bigforall}{\mathop{\mbox{\Large $\forall$}}}

\[\bigwedge_{i \in R} \bigg(\bigwedge_{j \in I} F_{ij}(H_j) \Rightarrow \bigwedge_{k \in O} F_{ik}(H_k)\bigg)\]
Here, $R$ is the set of all rows, $I$ is the set of all input columns, and $O$ is the set of all output columns.
As an example, applying this to the table in Figure \ref{fig:decisiontable} would produce the following formula: $(Age Of Person \geq 18 \Rightarrow Person Is Adult = Yes) \land (Age Of Person < 18 \Rightarrow Person Is Adult = No)$.

Now we extend this semantics to take variables and quantification into account.
Our first step is to define a function that maps cDMN expressions to terms. 
For the most part, this definition corresponds to that of Calvanese \cite{calvanese2017semantic}.
However, we extend it to take into account the fact that certain expressions -- which we call {\em variable expressions} -- must be translated to FO variables. 
There are three kinds of variable expressions. 
We now define a mapping $\nu$ that maps each of these three kinds of cDMN variable expressions to a typed FO variable $x$ of type $T$, which we denote as $x[T]$:

\vspace{-0.5em}
\begin{itemize}
    \item The name $T$ of a type is a variable expression. We define $\nu(T) = x_{T}[T]$, with $x_{T}$ a new variable of type $T$.
    \item An expression $e$ of the form ``$Type$ called $v$'' is a variable expression. We define $\nu(e) = v[Type]$.
    \item If the header of a column contains an expression ``$Type$ called $v$'', then $v$ is a variable expression in all subsequent columns of the table and in its body. We define $\nu(v)$ as $v[Type]$.
\end{itemize}
\vspace{-0.5em}

Given this function $\nu$, we now define the following mapping $t_\nu(.)$ of cDMN expressions to terms.
\vspace{-0.5em}

\begin{itemize}
    \item For a constant $c$, $t_\nu(c) = c$; similarly, for an integer or floating point number $n$, $t_\nu(n) = n$;
    \item For an arithmetic expression $e$ of the form $e_1 \theta e_2$ with $\theta \in \{+,-,*,/ \}$, we define $t_\nu(e) = t_\nu(e_1) ~ \theta ~ t_\nu(e_2)$;
    \item For a variable expression $v$, we define $t_\nu(v) = \nu(v)$.
    \item For a function expression, i.e. ``$Function$ of $arg_1$ and ... and $arg_n$'': \\ ${t_\nu(X) = Function(t_\nu(arg_1),...., t_\nu(arg_n))}$ .
\end{itemize}

\vspace{-0.5em}
Similarly to Calvanese, we translate each entry $c$ in a cell $(i,j)$ of a table into a formula $F_{ij}(x)$ in one free variable x:
\vspace{-0.5em}

\begin{itemize}
    \item If $c$ is of the form "$\theta e$" with $\theta$ one of the relational operators $\{\leq, \geq, =, \neq\}$, then $F_{ij}(x)$ is the formula $x ~ \theta ~ t(e)$;
    \item If $c$ is of the form $Not~e$, then $F_{ij}$ is $x \neq t(e)$;
    \item If $c$ is a list $e_1, \ldots, e_n$, then $F_{ij}$ is $x = t(e_1) \lor \ldots \lor x = t(e_n)$. As a special case, if $c$ consists of a single expression e, then $F_{ij}$ is $x = t(e)$.
    \item If $c$ is a range, e.g. $[e_1, e_2)$, then $F_{ij}$ is $x \geq t(e_1) \land x < t(e_2)$.
\end{itemize}

We are now ready to define the semantics of a constraint table. If I is the set of input columns of the table, $O$ the set of output columns and $V \subseteq I$ the set of variable introducing columns, we define the semantics of the table as:

\vspace{-0.5em}
\[\bigforall_{l \in V} \nu(H_l):  \bigwedge_{i \in R} \bigg(\bigwedge_{j \in I} F_{ij}\big(t_\nu(H_j)\big) \Rightarrow  \bigwedge_{k \in O} F_{ik}\big(t_\nu(H_k)\big)\bigg)\]

For example, in Figure \ref{tab:agatha}, ${\nu(H_1) =  { x[ Person] }}$ and ${t_\nu(H_1) = x}$, \\ $ {t_\nu(H_2) = Hatee(t(H_1)) = Hatee(x)}$, which leads to the formula: \[\forall x[Person]: Hatee(x) < 3.\]


The above transformation turns each decision or constraint table $T$ into an FO(.) formula $\phi_T$. The glossary
and data tables together define a structure $S$ for part of the vocabulary. The domain of $S$ consists of the union
of the interpretations $I_t$ of all the types $t$. If $t$ is enumerated in the glossary, then $I_t$ is this enumeration.
Otherwise, $I_t$ consists of all the values that appear in a data table in a column of type $t$. The structure $S$
interprets all the relations / functions for which a data table is provided, and it interprets them by the set of tuples /
the mapping that is given in this table.

The set of “solutions’’ of a cDMN model is the set $MX(\Phi, S)$ of all model expansions of the structure $S$
w.r.t.~the theory $\Phi = \{ \phi_T \mid T$ is a constraint or decision table$\}$, i.e., the set of all structures $S’
\models \Phi$ that extend $S$ to the entire vocabulary.

\section{Implementation} \label{sec:implementation}

This section gives a brief overview of the inner workings of the \DMNp solver\footnote{\url{https://gitlab.com/EAVISE/cdmn/cdmn-solver}}.
The solver consists of two parts: a constraint solver (the IDP system), and a converter from \DMNp to IDP input.
In principle, any constraint solver could be used, but we chose the IDP system because of its flexibility.


The cDMN to IDP converter is built using Python3, and works in a two-step process.
It first interprets all tables in a \texttt{.xslx} sheet and converts them into Python objects.
For example, the converter parses all the glossary tables and converts them into a single \texttt{Glossary} object, which then creates \texttt{Type} and \texttt{Predicate} objects.
The created Python objects are then converted into IDP blocks.
More detailed information about this conversion can be found in the cDMN documentation\footnote{\url{https://cdmn.readthedocs.io/en/latest/index.html}}, along with an explanation of the usage of the solver and concrete examples of cDMN implementations.

\section{Results and discussion} \label{sec:results}

In this section we first look at three of the DM Community challenges, each showcasing a feature of \DMNp.
For each challenge, we compare the DMN implementations from the DM Community website with our own implementation in \DMNp.
Afterwards, we compare all challenges on size and quality.



\vspace{-1em}
\subsection{Constraint tables}
\begin{figure}
    \vspace{-2em}
    \includegraphics[width=\linewidth]{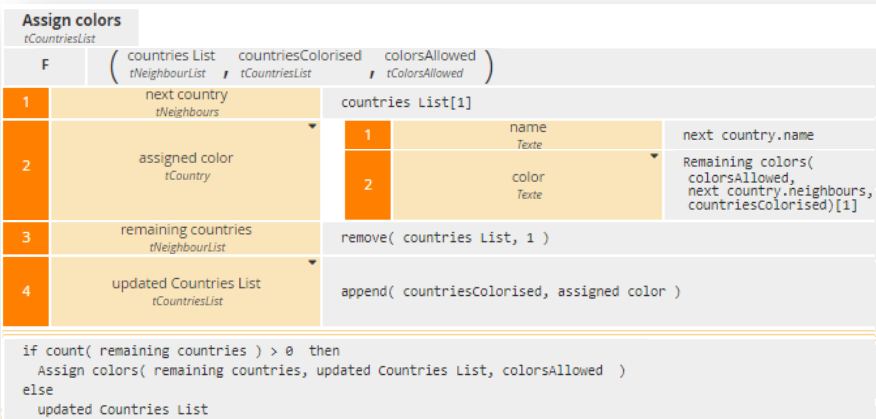}
    \vspace{-1.5em}

    \caption{An extract of the map coloring solution in standard DMN with FEEL}

    \label{fig:FEELmap}
    \vspace{-0.3cm}
\end{figure}
Constraint tables allow cDMN to model constraint satisfaction problems in a straightforward way.
For example, in \chalname{Map Coloring}, a map of six European countries  must be colored in such a way that no neighbouring countries share the same color.
For this challenge, a pure DMN implementation was submitted, of which Figure \ref{fig:FEELmap} shows an extract.
The implementation uses complicated FEEL statements to solve the challenge. 
While these statements are DMN-compliant, they are nearly impossible for a business user to write without help.
In \DMNp, we can use a single straight-forward constraint table to solve this problem, as shown in Figure \ref{tab:map}.
Together with the glossary and a data table (Figure \ref{tab:datatable_example}), this forms a complete yet simple \DMNp implementation.

\subsection{Quantification}
Quantification is useful in the \chalname{Monkey Business} challenge.
In this challenge, we want to know for four monkeys what their favorite fruit and their favorite resting place is, based on some information.
There are two DMN-like submissions for this challenge: one using Corticon, and one using OpenRules.

\begin{figure}
\vspace{-2em}
    \centering
    \begin{subfigure}[t]{\textwidth}
        \centering\includegraphics[width=8cm]{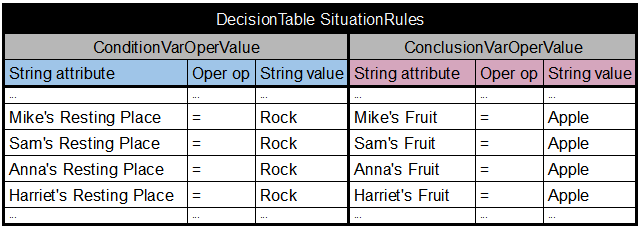}
        \caption{Open Rules}
        \label{fig:ormonkey}
    \end{subfigure}
    
    \begin{subfigure}[c]{\textwidth}
    {
        \dmnfont
        \centering
        \begin{tabular}{|r|l|l|l|l|l|}
            \cmidrule{1-3}
            \multicolumn{3}{|l|}{Monkey Constraints} & \multicolumn{1}{c}{}\\
            \hline
            E* & \inputcol \textbf{Monkey} & \inputcol \textbf{Place of Monkey} 
                 & \outputcol \textbf{Fruit of Monkey}\\
            \hline
            $\ldots$ & $\ldots$ & $\ldots$ & $\ldots$ \\
            \hline
            2 & --- & Rock &  Apple \\
            \hline
            $\ldots$ & $\ldots$ & $\ldots$ & $\ldots$ \\
            \hline
        \end{tabular}
        \caption{cDMN}
        \label{tab:monkey_business2}
    }
    \end{subfigure}
    \vspace{-1em}
    \caption{An extract of \chalname{Monkey Business} implementation in (a) OpenRules and (b) cDMN, specifying ``The monkey who sits on the rock is eating the apple''.}
    \vspace{-1em}
\end{figure}

    



        

One of the pieces of information is: \texttt{The monkey who sat on the rock ate the apple.}
The OpenRules implementation has a table with a row for each monkey, which states that if this monkey's resting place was a rock, their fruit was an apple (Figure \ref{fig:ormonkey}).
In other words, for $n$ monkeys, the OpenRules implementation of this rule requires $n$ lines.
Because of quantification, \DMNp requires only one row, regardless of how many monkeys there are (Figure \ref{tab:monkey_business2}).
The Corticon implementation also uses a similar quantification for this rule.

\begin{figure}
\vspace{-1em}
    \begin{subfigure}[c]{\linewidth}
        \centering
        \includegraphics[width=8cm]{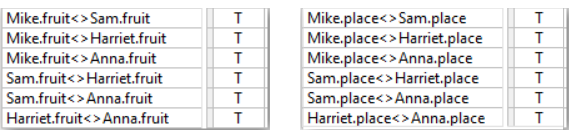}
        \vspace{-0.5em}
        \caption{Corticon}
        \label{fig:cormonkey}
    \end{subfigure}
    
    \begin{subfigure}[c]{\linewidth}
    \centering
    {
    \dmnfont
    \begin{tabular}{|r|l|l|l|l|}
        \cmidrule{1-3}
        \multicolumn{3}{|l|}{Different Preferences} & \multicolumn{2}{c}{}\\
        \hline
        E* & \inputcol \textbf{Monkey called m1} & \inputcol \textbf{Monkey called m2}
            & \outputcol \textbf{Place of m1} & \outputcol \textbf{Fruit of m1}\\
        \hline
        1 & --- & not(m1) & not(Place of m2) & not(Fruit of m2) \\
        \hline
    \end{tabular}
    
    \caption{cDMN}
    \label{tab:monkey_business1}
    }
    \end{subfigure}
    
    \vspace{-1em}
    \caption{An extract of the \chalname{Monkey Business} implementation in (a) Corticon and (b)c DMN, defining that no monkeys share fruit and no monkeys share the same place.}
    \vspace{-1em}
\end{figure}

Another rule states that no two monkeys can have the same resting place or fruit.
In both the Corticon and OpenRules implementations, this is handled by two tables with a row for each pair of monkeys.
The Corticon tables are shown in Figure \ref{fig:cormonkey}.
Each row either states that two monkeys have different fruit, or that they have different place.
Therefore, $n$ monkeys require $ \frac{n!}{(n/2)! * 2^{(n/2)}} $ rows.
By contrast, the \DMNp implementation seen in Figure \ref{tab:monkey_business1} requires only a single row to express the same.

\subsection{Optimization}
In the \chalname{Balanced Assignment} challenge, 210 employees need to be divided into 12 groups, so that every group is as diverse as possible.
The department, location, gender and title of each employee is known.
This is quite a complex problem to handle in DMN.
As such, of the four submitted solutions, only one was DMN-like: an OpenRules implementation, using external CP/LP solvers.
The logic for these external solvers is written in Java.
Although the code is fairly compact, it cannot be written without prior programming knowledge.
Because optimization is built-in in cDMN, we can solve the problem with two decision tables and one constraint table. The table \textit{Diversity score}, shown in Figure \ref{tab:balanced_assignment}, adds 1 to the total diversity score if two similar people are in a different group. Minimizing this score then results in the most diverse groups.

\begin{figure}[h]
    \vspace{-1.5em}
    \centering
    \dmnfont
    {
    \begin{tabular}{|r|p{1.2cm}|p{1.2cm}|p{1.6cm}|p{1.2cm}|p{1.1cm}|p{0.75cm}|l|l|}
        \cmidrule{1-8}
        \multicolumn{8}{|l|}{Diversity score} & \multicolumn{1}{c}{}\\
        \hline
        C+ & \inputcol \textbf{Person called p1} & \inputcol \textbf{Person called p2}
            & \inputcol \textbf{Department of p1} & \inputcol \textbf{Location of p1}
            & \inputcol \textbf{Gender of p1} & \inputcol \textbf{Title of p1}
            & \inputcol \textbf{Group of p1} & \outputcol \textbf{Score}\\
        \hline
        1 & - & - & = Department of p2 & - & - & - & not(Group of p2) & 1 \\
        \hline
        2 & - & - & - & = Location of p2 & - & - & not(Group of p2) & 1 \\
        \hline
        3 & - & - & - & - & = Gender of p2 & - & not(Group of p2) & 1 \\
        \hline
        4 & - & - & - & - & - & = Title of p2 & not(Group of p2) & 1 \\
        \hline
    \end{tabular}
    
    \begin{tabular}{|l|}
        \cmidrule{1-1}
        \multicolumn{1}{|l|}{\executecol \textbf{Execute}}\\
        \hline
        Minimize Score \\
        \hline
    \end{tabular}
    }
    \vspace{-1em}
    \caption{The decision tables and constraint table for Balanced Assignment.}
    \label{tab:balanced_assignment}
    \vspace{-4em}
\end{figure}

\subsection{Overview of all challenges}

Of the 21 challenges we considered, \DMNp is capable of successfully modeling 19.
In comparison, there were 12 OpenRules implementations and 12 Corticon implementations submitted.
Note that we have not examined whether OpenRules and Corticon might be capable of modeling more challenges than those for which a solution was submitted.

To compare \DMNp to other approaches, we focus on two aspects: quantity (how big are they?) and quality (how readable and how scalable are they?).
The size of implementations was measured by counting the number of cells used in all the decision tables. 
Glossaries, data tables and equivalents thereof were not included in the count.
Table \ref{table:cellcount} shows that \DMNp and Corticon alternate between having the fewest cells, and that OpenRules usually has the most.
In general, OpenRules implementations require many cells because each cell is very simple.
For instance, even an ``='' operator is its own cell.
The Corticon implementations, on the other hand, contain more complex cells, rendering them more compact.

\renewcommand*\rot{\multicolumn{1}{R{90}{1em}}}
\begin{table}
    \vspace{-0.3cm}
    
    \centering
    \begin{tabular}{c |c |c |c |c |c |c |c |c |c |c |c |c |c |c |c |c |c |c |c |}

        & \rot{Who Killed Agatha?}
        & \rot{Change Making Decision }
        & \rot{Make a Good Burger}
        & \rot{Duplicate Product Lines}
        & \rot{Collection of Cars}
        & \rot{Monkey Business}
        & \rot{Vacation Days}
        & \rot{Family Riddle}
        & \rot{Greeting a Customer}
        & \rot{Online Dating Services}
        & \rot{Classify Employees}
        & \rot{Reindeer Ordering}
        & \rot{Zoo, Buses and Kids}
        & \rot{Balanced Assignment}
        & \rot{Vacation Days Advanced}
        & \rot{Map Coloring}
        & \rot{Map Coloring Violations}
        & \rot{Crack the Code}
        & \rot{Numerical Haiku}
        \\
        \hline
        \DMNp   & 53\firstval& 26\secondval & 35\secondval & 20\secondval & 26\firstval & 47\firstval & 38\fourthval & 76\secondval & 88\firstval & 45\firstval & 36\thirdval & 14\firstval & 24\firstval & 55\secondval & 124\secondval & 21\firstval & 48\firstval & 77\firstval & 41\firstval \\
        \hline 
        Corticon & 54\secondval &  14\firstval & 20\firstval & 19\firstval  & 45\secondval & 64\secondval & 32\thirdval &  22\firstval & \emptyval &  78\secondval &  21\firstval & 64\secondval & \emptyval & \emptyval & \emptyval & \emptyval & \emptyval & \emptyval & \emptyval\\
        \hline
        OpenRules & 176\thirdval & \emptyval & 95\fourthval &  21\thirdval & \emptyval &  150\thirdval & 31\secondval & \emptyval &  205\secondval & \emptyval &  70\fourthval & 111\thirdval &  43\secondval & 30\firstval & 97\firstval & \emptyval & \emptyval & \emptyval & \emptyval\\
        \hline
        Others & \emptyval & \emptyval & 76$^1$\thirdval & \emptyval & 48$^1$\fourthval & \emptyval & 14$^2$\firstval & \emptyval & \emptyval & \emptyval & 34$^3$\thirdval & 370$^4$\fourthval &\emptyval &\emptyval &\emptyval & 31$^4$\secondval & \emptyval& \emptyval& \emptyval \\
        \hline
        
    \end{tabular}
    \caption{Comparison of the number of cells used per implementation. Other implementations: 1. FEEL, 2. Blueriq, 3. Trisotech, 4. DMN}
    \label{table:cellcount}
    \vspace{-1.5em}
\end{table}
\renewcommand*\rot{\multicolumn{1}{R{0}{1em}}}

\begin{figure}[h!]
    \centering
    \begin{subfigure}[c]{\linewidth}
    \centering\includegraphics[width=8cm]{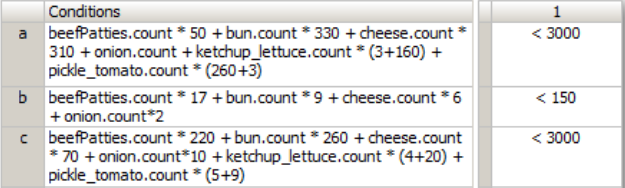}
    \caption{Corticon}
    \label{fig:hamburger_comparison}
    \end{subfigure}

    \begin{subfigure}[c]{\linewidth}
    {
    \centering
        \dmnfont
        \begin{tabular}{|r|l|p{1.9cm}|p{1.9cm}|p{1.9cm}|p{1.9cm}|}
            \cmidrule{1-2}
            \multicolumn{2}{|l|}{Determine Nutrition} & \multicolumn{4}{c}{}\\
            \hline
            C+ & \inputcol \textbf{Item} & \outputcol \textbf{Total Sodium}
                & \outputcol \textbf{Total Fat} & \outputcol \textbf{Total Calories}
                & \outputcol \textbf{Total Cost}\\
            \hline
            1 & - & Number of Item \newline * Sodium of Item & Number of Item \newline * Fat of Item & Number of Item \newline * Calories of Item & Number of Item \newline * Cost of Item \\
            \hline
        \end{tabular}
        \begin{tabular}{|r|l|l|l|}
            \cmidrule{1-1}
            \multicolumn{1}{|l|}{Nutrition Constraints} & \multicolumn{3}{c}{}\\
            \hline
            E* & \outputcol \textbf{Total Sodium}
                & \outputcol \textbf{Total Fat} & \outputcol \textbf{Total Calories}\\
            \hline
            1 & \textless 3000 & \textless 150 & \textless 3000  \\
            \hline
        \end{tabular}
        \caption{cDMN}
        \label{tab:hamburger_comparison}
    }
    \end{subfigure}
    \vspace{-0.5em}
    \caption{Calculating the food properties of burger in Corticon and cDMN.}
    \vspace{0em}
\end{figure}

Because of this, OpenRules implementations are usually easier to read than their Corticon counterparts.
An example comparison between \DMNp and Corticon can be seen in Figure \ref{fig:hamburger_comparison} and \ref{tab:hamburger_comparison}. 
Each figure shows a snippet of their \chalname{Make a Good Burger} implementation, in which the food properties of a burger are calculated.
While the Corticon implementation is more compact, it is less interpretable, less maintainable and dependent on domain size.
If the user wants to add an ingredient to the burger, complex cells need to be changed.
In \DMNp, simply adding the ingredient to the data table suffices.

\begin{figure}[H]
    \vspace{-1em}
    \begin{subfigure}[c]{\linewidth}
    \centering
    \includegraphics[width=8cm]{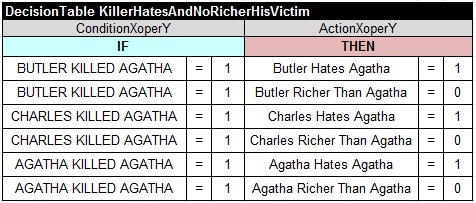}
    \caption{OpenRules}
    \label{fig:agatha_comparison}
    \end{subfigure}

    \begin{subfigure}[c]{\linewidth}
    {
    \centering
    \dmnfont
    \begin{tabular}{|r|l|l|l|}
        \cmidrule{1-1}
        \multicolumn{1}{|l|}{Killer constraints} & \multicolumn{2}{c}{}\\
        \hline
        E*  & \outputcol\textbf{Killer hates Agatha} & \outputcol\textbf{Killer richer than Agatha}\\
        \hline
        1  & Yes & No \\
        \hline
    \end{tabular}
    \caption{cDMN}
    \label{tab:agatha_comparison}

    }
    \end{subfigure}

    \caption{Implementation of ``A killer always hates and is no richer than their victim'' in OpenRules and cDMN}
    \vspace{-2em}
\end{figure}

A comparison between \DMNp and OpenRules can be found in Figure \ref{fig:agatha_comparison} and \ref{tab:agatha_comparison}.
Here we show a part of their implementations of the \chalname{Who Killed Agatha?} challenge.
They both show a translation of the following rule: ``A killer always hates, and is no richer than his victim.''
By using constraints and a constant (\texttt{Killer}), \DMNp allows us to form a more readable and more scalable table.
If the police ever find a fourth suspect, they can easily add the person to the data table without needing to change anything else.

In Section \ref{sec:chaloverview}, we identified four different problem properties.
We now suggest that each property is tackled more easily by one or more of the additions \DMNp proposes.

\vspace{-1em}
\subsubsection*{Aggregates needed}
Figure \ref{tab:hamburger_comparison} shows how aggregates are both more readable and scalable when using quantification.
Moreover, \DMNp allows the use of aggregates for more complex operations such as optimization or defining constraints.

\vspace{-1em}
\subsubsection*{Constraints}
Constraints can be conveniently modeled by constraint tables, such as the constraints in Figure \ref{tab:agatha_comparison}, which states that the killer hates Agatha, but is no richer than her.
The addition of constraint tables allows for an obvious translation from the rule in natural language to the table.

\vspace{-1em}
\subsubsection*{Universal quantification}
Problems which contain universal quantification can be compactly represented, as can, among others, be seen in Figure \ref{tab:agatha}. This table states that each person hates less than 3 people.

\vspace{-1em}
\subsubsection*{Optimization}
Because \DMNp directly supports optimization, problems containing this property are easily modeled.
Furthermore, by the addition of more complex data types, optimization can be used in a more flexible manner.
An example can be found in Figure \ref{tab:balanced_assignment}.
A summary of each problem property and its \DMNp answer can be found in Table \ref{table:propsol}.

\begin{table}[H]

    \centering
    \begin{tabular}{l | l}
        \textbf{Property} & \textbf{cDMN answer}\\
        \hline
        Aggregates needed & Quantification, expressive data\\
        Constraints & Constraint tables, quantification, expressive data\\
        Universal quantification & Quantification \\
        Optimization & Optimization, expressive data\\
    \end{tabular}
    \caption{A comparison between the problem properties and their cDMN answers.}
    \label{table:propsol}

\end{table}

\section{Conclusions} \label{sec:conclusion}

This paper presents an extension to DMN, which is able to solve complex problems while still maintaining DMN's level of readability.
This extension, which we call \DMNp, adds constraint modeling, more expressive data and quantification.

Constraint modeling allows a user to define a solution space instead of a single solution.
A user can generate a desired number of models, or generate the model which optimizes the value of a specific term.
Unlike DMN, which only knows constants, \DMNp also supports the use of functions and predicates, which allow for more flexible representations.
Together with quantification, this allows tables to be constructed in a compact and straightforward manner, while being independent of the size of the problem.
This improves readability, maintainability and scalability of tables.

By comparing our \DMNp implementations to the implementations of other state-of-the-art DMN-like solvers, we can conclude that \DMNp succeeds in increasing expressivity while retaining the simplicity of standard DMN.
\printbibliography
\end{document}